%% file: transformer_hls4ml.tex
\documentclass[conference]{IEEEtran}
\IEEEoverridecommandlockouts
\usepackage[numbers,sort,compress]{natbib}

\usepackage[utf8]{inputenc} %
\usepackage[T1]{fontenc}    %
\usepackage{hyperref}       %
\usepackage{url}            %
\usepackage{booktabs}       %
\usepackage{amsfonts}       %
\usepackage{nicefrac}       %
\usepackage{microtype}      %
\usepackage{lipsum}
\usepackage{graphicx}
\usepackage{subcaption}
\usepackage{xcolor}
\usepackage{multirow}
\graphicspath{ {./figures/} }
\usepackage{xspace}
\usepackage{amsmath}
\usepackage{doi}

\begin{document}
 
\title{Low Latency Transformer Inference on FPGAs for Physics Applications with hls4ml}

\author{
\IEEEauthorblockN{Zhixing Jiang, Dennis Yin, Yihui Chen,
\\ Elham E Khoda, Scott Hauck, Shih-Chieh Hsu,}
\IEEEauthorblockA{\textit{University of Washington}\\
Seattle, WA 98195, USA}
\and
\IEEEauthorblockN{Ekaterina Govorkova, Philip Harris, \\
Vladimir Loncar, Eric A. Moreno, }
\IEEEauthorblockA{\textit{Massachusetts Institute of Technology}\\
Cambridge, MA 02139, USA}
\and

}

\maketitle
\begin{abstract}
This study presents an efficient implementation of transformer architectures in Field-Programmable Gate Arrays (FPGAs) using \texttt{hls4ml}.
We demonstrate the strategy for implementing the multi head attention, softmax, and normalization layer and evaluate three distinct models.
Their deployment on VU13P FPGA chip achieved latency less than 2 $\mu$s, demonstrating the potential for real-time applications.
\texttt{hls4ml}'s compatibility with any TensorFlow-built transformer model further enhances the scalability and applicability of this work.

\end{abstract}


\input{01_intro}

\input{02_background}

\input{03_relatedwork}
\input{04_implementation}
\input{05_benchmark_studies}

\input{06_performance}

\input{07_conclusion}

\section*{Acknowledgements}

We acknowledge the Fast Machine Learning collective as an open community of multi-domain experts and collaborators. This community was important for the development of this project. We also wish to thank the A3D3 collaboration for their guidance and support, which played a crucial role in shaping our work.

\bibliographystyle{IEEEtran}
\bibliography{references}

\end{document}

%% file: 01_intro.tex
\section{Introduction}
\label{sec:intro}

Over the past few years, machine learning (ML) has firmly established itself as an indispensable tool in diverse scientific and industrial fields, pushing the boundaries of innovation and productivity to unprecedented levels. Among the rich suite of ML techniques available, the transformer architecture has demonstrated exceptional prowess in handling a diverse array of complex problems. While its initial fame was earned in the domain of natural language processing (NLP), its utility has since transcended these bounds, proving to be also efficient in signal processing tasks. Examples of these include processing data from the LHC~\cite{LHC_2008}, detecting gravitational waves~\cite{Aasi_2015, Acernese_2015, KAGRA:2020tym}, among many others.

The transformer model, introduced by Vaswani et al.~\cite{NIPS2017_3f5ee243}, has revolutionized how we process sequence data. In contrast to traditional sequence-processing models, which sequentially process input data, transformers employ a unique mechanism of attention, allowing them to process data components independently and in parallel. In particular, the transformer employs a mechanism called the multi-head self-attention mechanism. This mechanism calculates the relationships or relevance scores among all pairs of data components in a sequence, thereby effectively encoding contextual information. With this design, transformers can capture long-range dependencies in sequence data, a feature particularly beneficial for complex, multidimensional processing tasks.

Transformer models, typically run on Graphics Processing Units (GPUs) and Central Processing Units (CPUs), encounter latency issues during real-time data processing that inhibit their instant response capability. This study suggests deploying transformers on Field-Programmable Gate Arrays (FPGAs) to leverage their adaptability, energy efficiency, and parallel processing capacity, which align with the demands of transformer computations. Recent work on deploying transformers on FPGA includes~\cite{Li_2020_ISLPED, Wojcicki_2022_ICFPT, Tzanos_2022_PACET, Han_2023_FPL}. We propose an automatic conversion approach using the \texttt{hls4ml} compiler, an high-level synthesis (HLS) based tool~\cite{Duarte:2018ite}, that supports various neural networks~\cite{CNN_hls4ml, RNN_hls4ml, RNN_GW_hls4ml, GNN_hls4ml}. We have extended \texttt{hls4ml} to transform any TensorFlow-based transformer model into an FPGA-friendly format, enhancing the versatility of this method.

Deploying transformers on FPGAs unlocks the potential for real-time, resource-efficient processing in various fields, extending beyond high-energy physics to encompass gravitational wave detection and automotive anomaly recognition. The versatility of FPGAs is demonstrated in our benchmarks, which include processing the high-frequency experiment of the LHC, as well as discerning intricate gravitational wave signals and pinpointing anomalies in car engine operation. In each scenario, the criticality of managing vast data streams at high speeds is paramount. To address the data deluge and expedite analysis, these applications employ a refined online selection system, executing on hardware like ASICs or FPGAs to efficiently process and triage data.

\begin{figure}[!ht]
\centering
\includegraphics[width=0.48\textwidth]{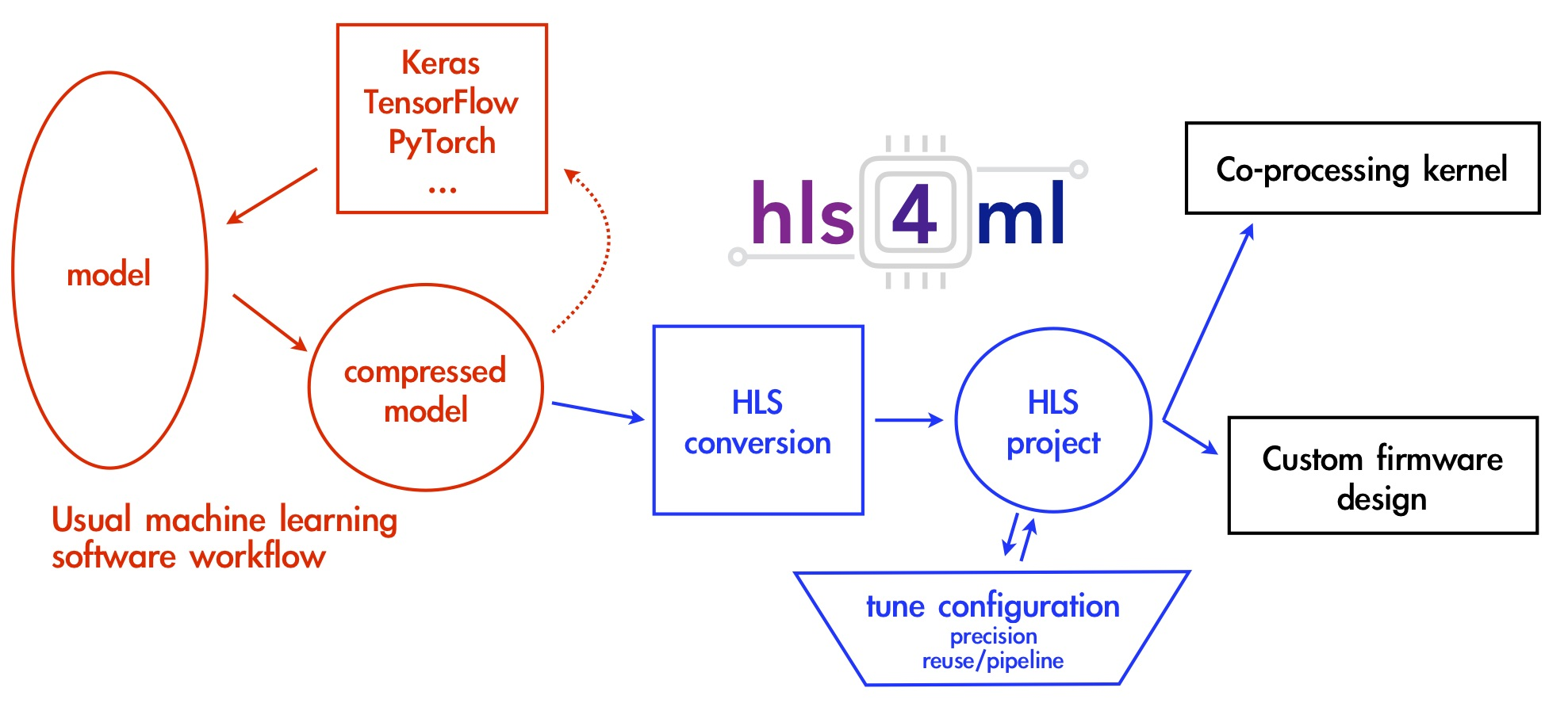}
\caption{The workflow of \texttt{hls4ml}~\cite{Duarte:2018ite}}
\label{fig:hls4ml}
\end{figure}

To handle the myriad of relevant tasks in diverse areas, we introduce transformer support in \texttt{hls4ml}, which can efficiently convert any TensorFlow-built transformer model into an FPGA-compatible form. The implementation ensures efficient resource utilization, low-latency performance, and model compatibility, delivering an enhanced ML experience on FPGAs. The workflow of \texttt{hls4ml}, as shown in figure~\ref{fig:hls4ml}, is designed with the aim of abstracting the complexities of FPGA programming. The framework takes trained ML models, primarily from TensorFlow or PyTorch, and translates them into HLS code. This code can then be synthesized into digital circuits using HLS tools. \texttt{hls4ml} maintains a high degree of model compatibility, which is crucial in scenarios where the machine-learning model may require regular updates or alterations. By bringing transformers to FPGAs, we extend the benefits of transformer models to a range of low-latency applications, opening new possibilities for real-time ML processing in various domains.

The paper is organized as follows: Section~\ref{sec:background} introduces the background of the transformer architecture and the concept of using \texttt{hls4ml}. Section~\ref{sec:related} discusses related works. Section~\ref{sec:implementation_detail} examines the implementation details, providing an in-depth look at the process of implementing transformers. Section~\ref{sec:benchmark_studies} provides comprehensive benchmarks, including a car-engine anomaly detection task, a B-tagging classification task, and a gravitational waves (GW) classification task. Finally, Section~\ref{sec:performance} presents the performance, resource, and latency estimates.

%% file: 02_background.tex
\section{Background}
\label{sec:background}

\subsection{Detailed Description of Transformer Architecture}
\label{subsec:trans_arch}
\begin{figure}[!ht] 
    \centering
    \includegraphics[width=0.4\textwidth]{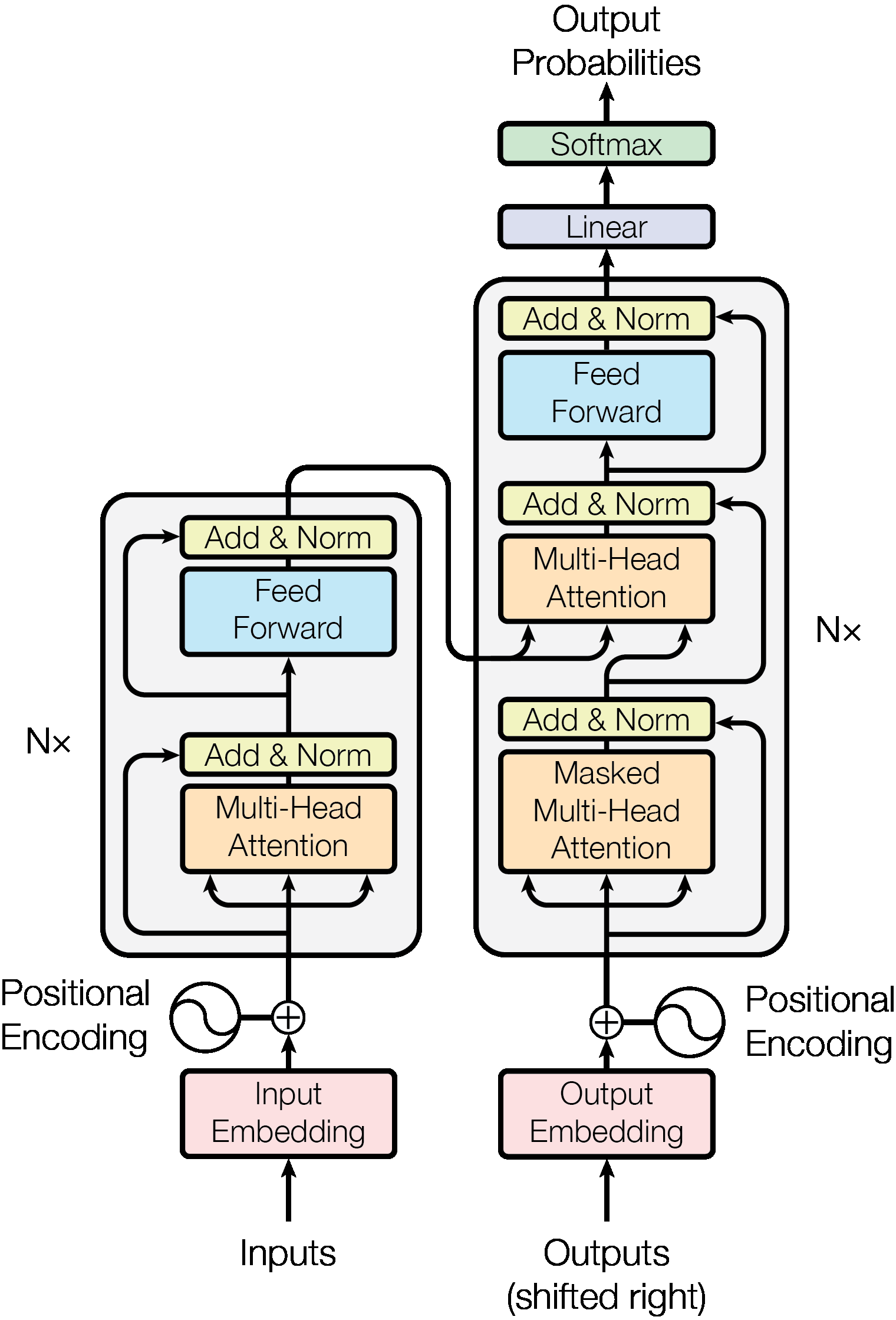} 
    \caption{The architecture of the transformer model~\cite{NIPS2017_3f5ee243}}
    \label{fig:transformer}
\end{figure}

\begin{figure*}[!ht] 
    \centering
    \includegraphics[width=1.0\textwidth]{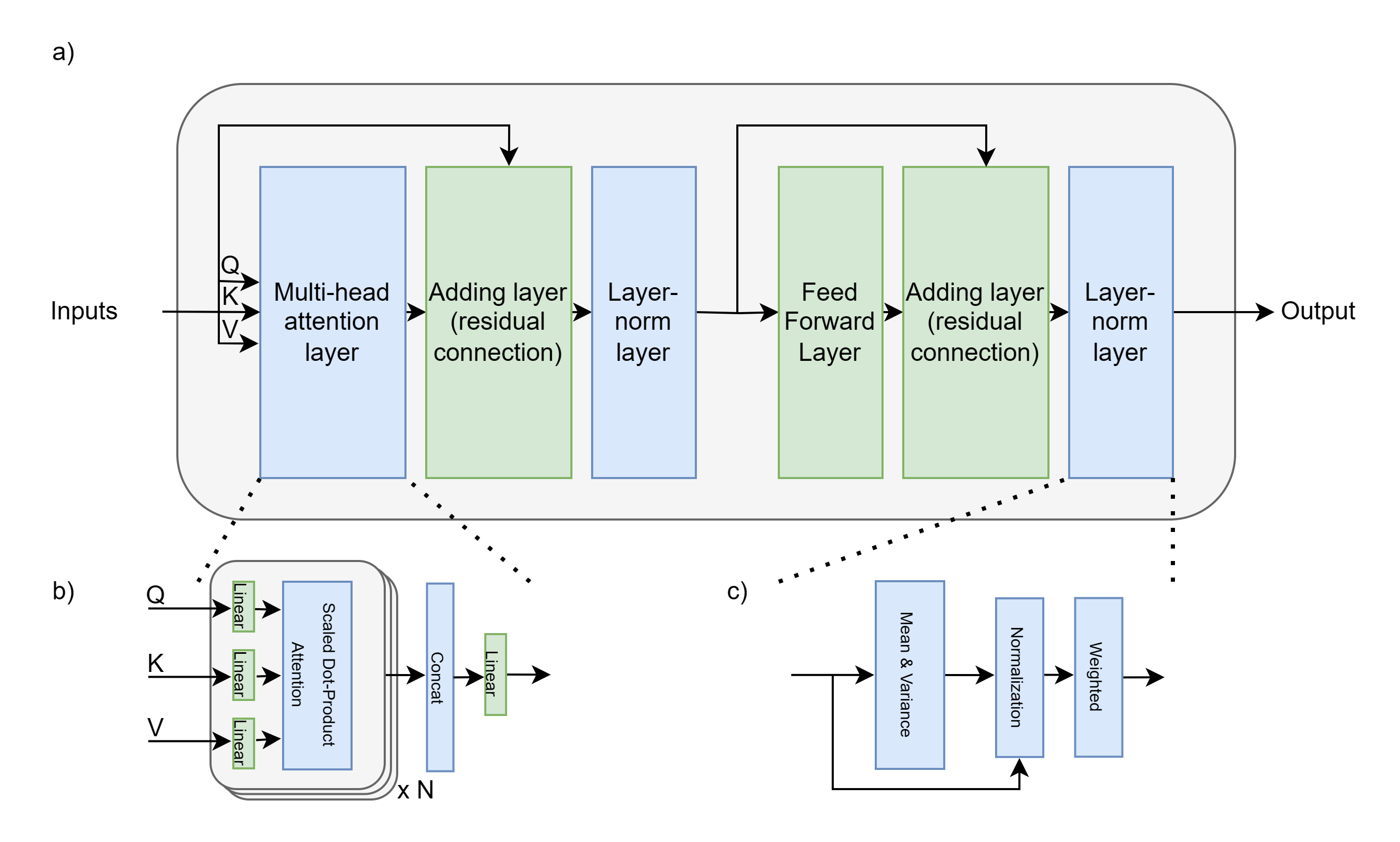} 
    \caption{One transformer block. The green layers are existing \texttt{hls4ml} functionality, while the blue are new in this paper.}
    \label{fig:transformer_block}
\end{figure*}

The transformer is a groundbreaking model introduced by Vaswani et al.~\cite{NIPS2017_3f5ee243} for processing sequence data. This architecture has had a significant impact, particularly in the field of natural language processing, while also demonstrating its utility in other areas, such as signal processing. The architecture of the transformer is shown in figure~\ref{fig:transformer}. A transformer model consists of several blocks, as shown in figure~\ref{fig:transformer_block}. Each block incorporates a multi-head self-attention (MHA) layer and a feed-forward neural network layer. These layers are interconnected through residual connections and are subsequently followed by layer normalization.

The MHA layer is the key to the transformer architecture. It has the ability to discern complex dependencies in the data, regardless of their position in the sequence. Within the MHA layer, there are multiple heads that compute distinct learned linear projections of the input data. The "multi-head" aspect of the attention mechanism enables the model to focus on different features in the data simultaneously, enhancing its ability to capture various types of information.

Each head within the MHA layer functions independently. For every head, three learnable matrices, known as query (\(Q\)), key (\(K\)), and value (\(V\)) matrices, are defined. The input vector is then linearly transformed by these matrices. For each head, these transformations are computed as follows:

\begin{gather}
\label{eq:grouped_eq}
Q = W_qX \\
K = W_kX \\
V = W_vX
\end{gather}

In the above equations, \(X\) is the input to the layer, and \(W_q\), \(W_k\), and \(W_v\) are trainable weight matrices. The relationships among all data in a sequence are computed by attending to the \(Q\), \(K\), and \(V\) matrices. This is achieved through the scaled dot-product attention mechanism, where the attention score between any two positions in a sequence is computed as the dot product of their corresponding \(Q\) and \(K\) vectors, scaled by the square root of their dimensionality, and then passed through a softmax function to obtain attention probabilities. Finally, these probabilities are used to form a weighted sum of the value (\(V\)) vectors. Each row in the output matrix (\(O_h\)) corresponds to the output for a particular position in the sequence, computed as a weighted sum of all value vectors, with the weights given by the attention probabilities. As described, the output, \(O_h\), for each head is computed as:

\begin{equation}
\label{eq:head_out}
O_h = \text{softmax}\left(\frac{QK^T}{\sqrt{d_k}}\right)V
\end{equation}

In the above function, \(d_k\) is the dimensionality of the key vectors. After obtaining \(O\) for each head (\(O_1\), \(O_2\), ..., \(O_n\)), the outputs from all heads are then concatenated and linearly transformed to yield the final output of the MHA layer:

\begin{equation}
O_{\text{final}} = \text{Concat}(O_1, O_2, ..., O_n)W_o
\end{equation}

where \(W_o\) is another learned weight matrix, and \(n\) is the number of heads.

This mechanism allows the transformer to allocate variable amounts of "attention", or importance, to different parts of the input sequence when processing data, leading to its outstanding performance in numerous tasks.

%% file: 03_relatedwork.tex
\section{Related Work}
\label{sec:related}

The study proposed by Wojcicki et al.~\cite{Wojcicki_2022_ICFPT} provided significant insights into implementing Transformer Neural Networks (TNNs) on FPGAs. The team from Imperial College London created a customized TNN architecture for FPGAs that notably outperformed GPU-based models in speed while maintaining comparable accuracy. They also proposed a novel, model-independent, post-training quantization search algorithm adaptable to various hardware environments. Besides the work from Wojcicki et al.~\cite{Wojcicki_2022_ICFPT}, other transformer implementations on FPGAs have been developed, such as those by Li et al.~\cite{Li_2020_ISLPED}, Peng et al.~\cite{Peng_ISQED_2021}, Tzanos et al.~\cite{Tzanos_2022_PACET}, Hong et al.~\cite{hong2022dfx}, and Han et al.~\cite{Han_2023_FPL}. However, these works mainly focus on the FPGA implementation of specific models. Our research extends this by developing an auto-converting mechanism for all transformer models generated by Keras, broadening the applicability and flexibility of transformer models on FPGAs.

%% file: 04_implementation.tex
\section{Implementation Detail}
\label{sec:implementation_detail}

This section outlines the specific implementation details of various components of the transformer architecture, focusing on the multi-head attention (MHA) layer, SoftMax layer, and normalization layer. These layers are implemented on FPGA hardware using the \texttt{hls4ml} tool, with each layer's implementation carefully optimized to deliver efficient resource utilization and performance.

\subsection{Multi-Head Attention Layer}
The MHA layer is an important component of the transformer architecture, and its efficient implementation plays a crucial role in the successful integration of transformer models into FPGAs. The MHA layer's operation was described in section~\ref{subsec:trans_arch}.

\begin{figure}[!ht] 
    \centering
    \includegraphics[width=0.52\textwidth]{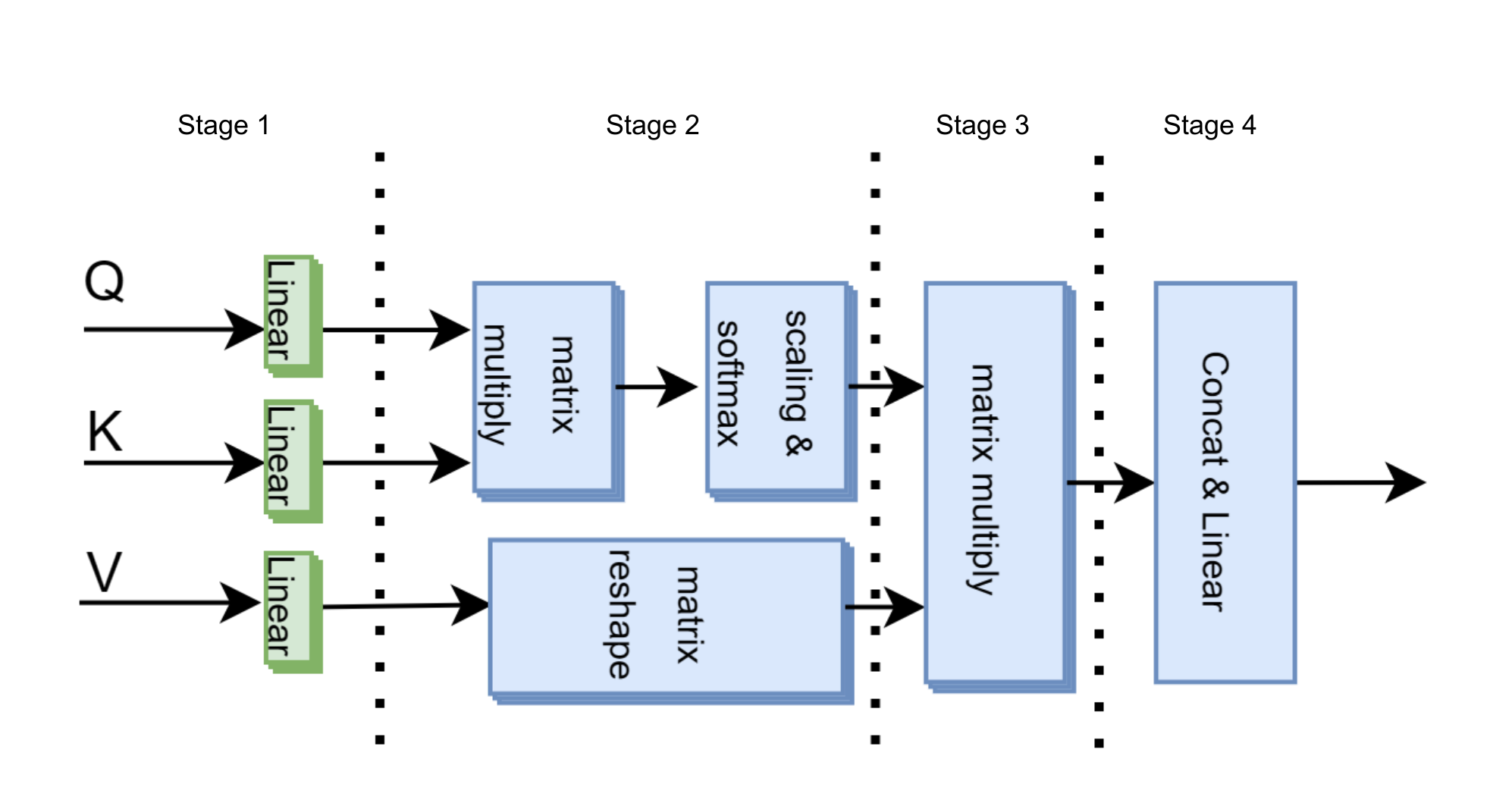} 
    \caption{The pipeline stages for the MHA layer}
    \label{fig:mha_imp}
\end{figure}

Implementing the MHA layer involves several complex operations, including linear projections, matrix multiplications, application of the SoftMax function, and final concatenation with another linear projection. To efficiently manage these operations, the implementation process is designed as four sequential pipeline stages, as shown in figure~\ref{fig:mha_imp}.

\underline{The first stage} of the MHA layer is the linear projection stage. This stage initiates the transformation process by transforming the original input sequence into three distinctive components - Query (\(Q\)), Key (\(K\)), and Value (\(V\)) vectors. These vectors are produced by applying separate weight matrices (\(W_Q\), \(W_K\), and \(W_V\)) to the input, as shown in the equation~\ref{eq:grouped_eq}.

\begin{figure}[!ht] 
    \centering
    \includegraphics[width=0.4\textwidth]{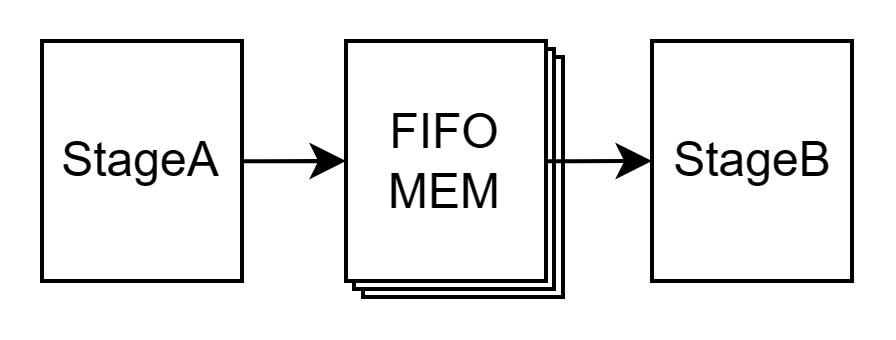} 
    \caption{The data streaming structure between layers using FIFO memory}
    \label{fig:fifo_mem}
\end{figure}

Inside this stage, a matrix times a vector operation is performed at each time step. In other words, the stage 1 itself is also a pipelined process, which would produce one row of the output matrix at each time step, with each row representing a time step of data. To optimize FPGA resources, vectors from this stage are stored in a FIFO memory structure as shown in the figure~\ref{fig:fifo_mem}. There are multiple FIFO memories stacking together in order to increase the bandwidth. The number of FIFO memories depends on the number of reuse factor (the concept of reuse factor is introduced in the section~\ref{subsec:Parallelization}) and the number of outputs of the previous layer.

\begin{figure}[!ht] 
    \centering
    \includegraphics[width=0.5\textwidth]{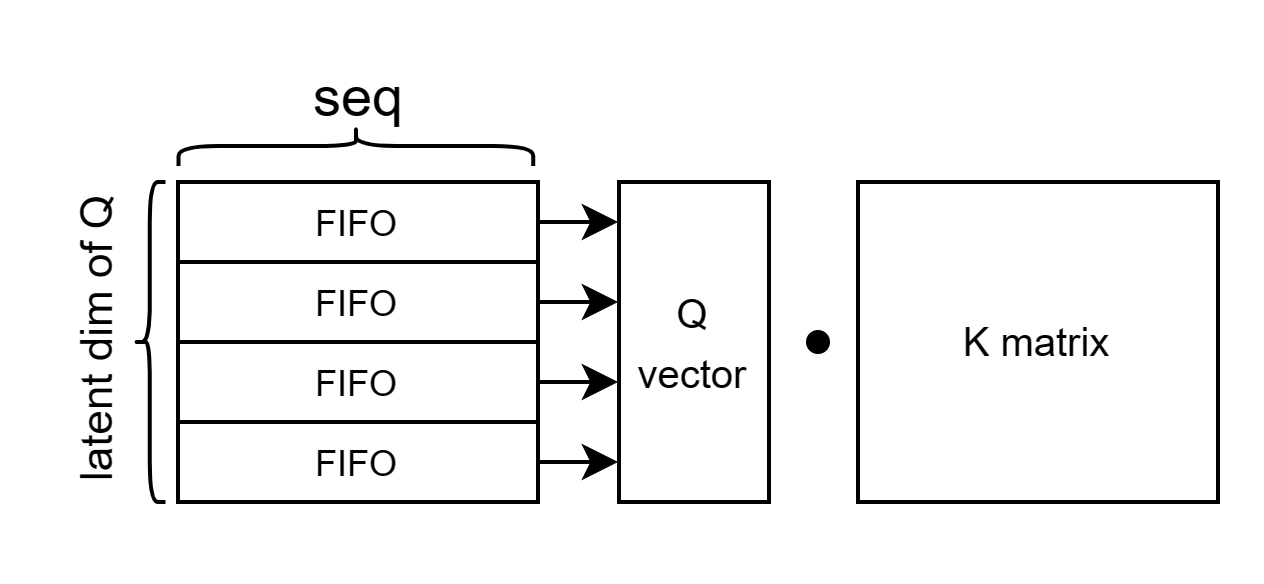} 
    \caption{The \(Q\) and \(K\) multiplication in hardware. \(Q\) is stored in FIFOs, while the \(K\) can be fully partitioned into the register.}
    \label{fig:qk_mult}
\end{figure}

\underline{The second stage} marks the beginning of the attention computation, involving the dot product between the \(Q\) and \(K\) matrices to produce the score matrix. Here, we perform a dot product between the \(Q\) vector and the \(K\) matrix, as shown in figure~\ref{fig:qk_mult}, calculating a relevance score for each pair of input data elements in the sequence. After matrix multiplication, we apply a scaling factor \(\sqrt{d_k}\), which is shown in equation~\ref{eq:head_out}, where \(\sqrt{d_k}\) is a pre-calculated constant for a given trained model. The resulting matrix of scores is then passed to the SoftMax function, which is implemented using a lookup table. The output of the SoftMax function is stored in a FIFO memory for later processing. The implementation of the SoftMax function is explained in section~\ref{subsec:SoftMax}.

Meanwhile, a matrix reshape operation is performed on the matrix \(V\). In the previous step, the \(V\) vectors were stored row-wise for fast data writing into memory, but in stage 3, we need to access the matrix \(V\) both column-wise and row-wise concurrently. Thus, in stage 2, we perform a matrix reshape, which enables the matrix \(V\) to become fully accessible. 

To implement this operation efficiently on FPGAs, careful data flow and resource management are crucial. To enable parallel data access, \(K\) vectors are loaded into a two-dimensional register, allowing simultaneous retrieval of all \(K\) data points. This design not only supports parallel computation but also provides flexibility by allowing users to adjust the partition factor to control the data flow. As matrix multiplication proceeds row-by-row, this configuration ensures that all \(K\) vectors are immediately available, thereby facilitating fast, parallel processing.
\vspace{1em}

\underline{The third stage} of the MHA layer implementation involves the second matrix multiplication. In this process, the scaled and SoftMax-applied relevance scores are multiplied by their corresponding \(V\) vectors. 
Similar to the \(K\) vector in the previous step, for this stage, the \(V\) vectors are stored in a fully accessible register. This choice of memory allows each vector in \(V\) to be accessed in parallel during the multiplication process, ensuring the efficient use of computational resources and boosting the speed of the operation.
Concurrently, one row of the relevance score matrix, now stored in the FIFO memory after the SoftMax application, is loaded. This row is then multiplied by the \(V\) matrix to generate a new row in the output attention vector. This process is repeated until all rows of the score matrix have been processed.
Essentially, this third stage works like a weighted sum operation, where the relevance scores weight the \(V\) vectors. The outcome is an attention vector that preserves important information from the input data while discarding less relevant details. The resultant attention vector is stored in an output FIFO memory, ready for the next stage of processing.
\vspace{1em}

\underline{The fourth stage} of the MHA layer implementation involves two key processes: the concatenation of the output from all attention heads and the subsequent linear transformation of the concatenated result.
Each attention head provides an output vector, stored in the FIFO memory after the second matrix multiplication. These output vectors are loaded row by row, aligning with the temporal sequencing of the data. Once loaded, the outputs are concatenated together to form a single, unified data stream. This process effectively amalgamates the attention vectors from all heads, capturing the diverse contextual insights each head has extracted from the input data.
Following concatenation, the data stream is passed through a linear layer. This transformation serves to map the concatenated output to the desired dimensionality, effectively forming the final output of the MHA layer. The linear layer is also pipelined; it reads one row of data and outputs one row at a time. This stage manages the output from all heads and efficiently generates the final output.

\vspace{1em}

\subsection{SoftMax Layer}
\label{subsec:SoftMax}
In the MHA layer, SoftMax computation translates the raw scores from the dot product of the Query (Q) and Key (K) matrices into probabilities, indicating the relevance of different data elements in the sequence. The original SoftMax formula in \texttt{hls4ml} was:

\[ S_i = \left( \sum_{j=1}^{k} e^{(z_j - z_i)} \right)^{-1} \]

where \( z_i \) and \( z_j \) are individual elements of the input sequence, and \( k \) represents the total number of elements. This formula required each SoftMax output \( S_i \) to calculate the exponent of the difference between \( z_j \) and \( z_i \), sum these values across all elements, and then invert the result. This process had to be repeated for each element, leading to a total of \( k^2 \) operations, which was computationally intensive.

In response to this complexity, the SoftMax layer has been restructured to use a simpler formula:

\[ S_i = \left( \sum_{j=1}^{k} e^{z_j} \right)^{-1} \times e^{z_i} \]

\begin{figure}[!ht] 
    \centering
    \includegraphics[width=0.45\textwidth]{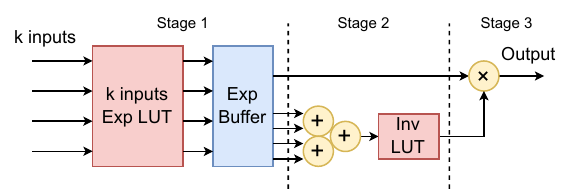} 
    \caption{The pipeline stages of the custom SoftMax layer}
    \label{fig:softmax_imp}
\end{figure}

This restructured formula was implemented through a three-step process, as shown in figure~\ref{fig:softmax_imp}, reducing computational overhead while maintaining the accuracy of the output.
\vspace{1em}

\underline{The first stage} is an element-wise exponentiation computation using the lookup table for all input values. This step yields a sequence of \( e^{z_j} \) values for each element in the sequence.
\vspace{1em}

\underline{The second stage} involves calculating the sum of the exponent values obtained in the first step. The sum is computed once for all elements in the sequence, inverted using an inversion lookup table, and the result stored in a register, which computed the value for \( (\sum_{j=1}^{k} e^{z_j})^{-1} \).

\vspace{1em}
\underline{The third stage} is an element-wise multiplication. The inverted sum from the register is multiplied element-wise with the \( e^{z_j} \) values for each element in the sequence. This process yields the SoftMax outputs for all elements.

\vspace{1em}
\noindent The revamped implementation of the SoftMax layer streamlines the computation, marking a significant improvement over the traditional approach. It considerably reduces the computation load and ensures an efficient and effective transformation of the MHA layer outputs into probability scores because it has a total of \( k \) operations instead of \( k^2 \).

\vspace{1em}
\subsection{Layer Normalization Layer}
\begin{figure}[!ht] 
    \centering
    \includegraphics[width=0.53\textwidth]{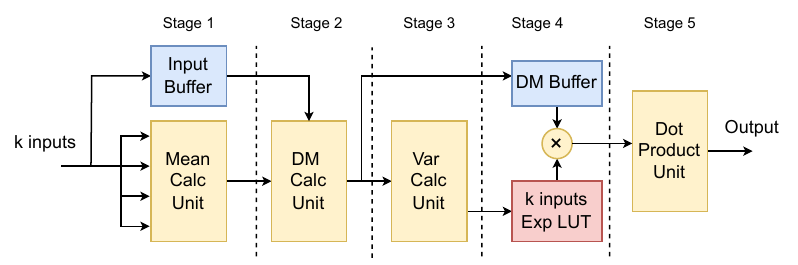} 
    \caption{The pipeline stages of the Layer Normalization layer}
    \label{fig:norm_imp}
\end{figure}

Layer normalization is crucial in the transformer model, as it standardizes the features of the input sequence across individual time steps. This process helps stabilize the neural network's learning and enhances the transformer's performance. Implementing the Layer Normalization layer involves multiple stages, each dedicated to a specific calculation within the normalization formula, as shown in figure~\ref{fig:norm_imp}.

\vspace{1em}
\underline{The first stage} calculates the mean value of the sequence for a given time step. This mean is obtained by summing all the elements \( x[j] \) for \( j \) in the range \( k \) and then multiplying by \( 1/k \), where \( k \) is the total number of elements.
\[ \text{mean} = \frac{\sum_{j=1}^{k} (x[j])}{k} \]

\underline{The second stage} computes the deviation from the mean (\( DM \)) for each element in the vector \( x \). The deviation, \( dm[j] \), for each element \( j \) in \( x \) is calculated as \( x[j] - \text{mean} \). 
\[ DM[j] = (x[j] - \text{mean}_i) \text{ for } j \text{ in range}(k) \]

\underline{The third stage} calculates the variance, which measures how spread out the elements in the sequence are. This is done by squaring the deviation of each element, summing them, and then multiplying by \(1/k \).
\[ \text{var} = \frac{\sum_{j=1}^{k} (DM[j]^2) }{k} \]

\underline{The fourth stage} is computing the normalized values. Each normalized value \( x_{i{\text{ normalized}}} \) is calculated by multiplying DM by the inverse of the square root of the variance. 
\[ x_{i{\text{ normalized}}} = \frac{DM}{\sqrt{\text{var}}} \]
The term \( \frac{1}{\sqrt{\text{var}}} \) is computed using a lookup table (LUT), enabling resource-efficient computation.

\underline{the fifth stage} calculates the final output. Each normalized value is scaled by a trainable parameter (\( \gamma \)) using a dot product unit, and a trainable offset (\( \beta \)) is added.
\[ \text{output}_i = x_{i{\text{ normalized}}} \times \gamma + \beta \]

%% file: 05_benchmark_studies.tex
\section{Benchmark Studies}
\label{sec:benchmark_studies}
 
In this Benchmark Studies section, we evaluate three distinct transformer models, each trained on a unique, complex dataset. These three datasets provide a view of potential applications of transformer models in FPGA deployments for different use cases. Our first model is a binary classifier that identifies anomalies in car engine data. This dataset has 3244 trainable parameters. The second model is a multi-class classifier trained on B-tagging data, a dataset obtained from CERN's Large Hadron Collider (LHC), which includes a larger parameter space with 9135 trainable parameters. Finally, our third model is based on the LIGO dataset, aiming to differentiate gravitational wave signals from background noise. This model handles 3394 parameters. All models are trained using TensorFlow and Keras. Details on the hyperparameters and specifications of each model can be found in TABLE~\ref{table:benchmark_spec}.

\begin{table}[h]
\centering
\caption{ Specifications of models}
\label{table:benchmark_spec}
\begin{tabular}{|l|l|l|l|}\hline 
Parameter & Engine & B-tagging & GW \\ \hline
Seq. Length & 50 & 15 & 100 \\ \hline
Input Vec. Size & 1 & 6 & 2 \\ \hline
No. of Transf. Blocks & 3 & 3 & 2 \\ \hline
Hidden Vec. Size & 16 & 64 & 32 \\ \hline
Output Vec. Size & 2 & 3 & 1 \\ \hline
Trainable Param. & 3244 & 9135 & 3394 \\ \hline
\end{tabular}
\end{table}

\subsection{Engine Anomaly Detection Model}
The first model is a binary classifier trained on a dataset that monitors and identifies anomalies within car engines. The FordA dataset we used is from the UCR/UEA archive~\cite{UCRArchive2018}, offering measurements collected from different engines in operation, with each engine's normal behavior considered as one class and any deviation from that normal behavior categorized as an anomaly, forming the second class. As the data is sourced directly from operating engines, it poses real-world challenges such as varying operational conditions, noise, and non-stationarity, making it a robust dataset for our model training.

As the first model in our series, we strived for a balance between simplicity and performance. We chose to forego the normalization layer for this model to maintain simplicity, but it does incorporate residual connections, which help prevent the vanishing gradient problem and facilitate deeper models. After the last MHA layer, the data enters two dense layers that further process the encoded representations. The final layer is a SoftMax layer, which provides a probabilistic distribution over the two classes: normal and anomalous. The SoftMax layer essentially classifies whether the instance falls into the normal or the anomalous category based on the values received from the preceding dense layers. Despite its simplicity, this model achieves an accuracy of 98\% on the validation set, showcasing its effectiveness.

\subsection{B-Tagging Model}
To benchmark our implementations, we study the open data samples from the Compact Muon Solenoid (CMS) experiment, which contain top quark pairs decaying hadronically with a center-of-mass energy of 7 TeV~\cite{ftag_dataset}. These events contain many bottom quark jets (b jets), charm quark jets (c jets), and jets from light quarks and gluons (light jets) originating from top quark decay.

Jets are collimated showers of particles that result from the decay and hadronization of quarks and gluons. At the LHC, an interesting jet signature emerges from overlapping quark-initiated showers produced in decays of heavy Standard Model particles like bottom quarks. The jets in the dataset are labeled as b, c, or light jets, depending on whether they contain bottom quarks, charm quarks, or neither, respectively. The task of identifying heavy jets like b jets from c jets and light jets is called flavor tagging. The main feature that separates b jets (and c jets) from light jets is the presence of the displaced vertex corresponding to the decay of the hadron containing the b (or c) quark. These hadrons are long-lived due to their mass, and the decay time depends on their momenta.

Our proposed algorithm aims to identify the presence of tracks consistent with these displaced vertices using a transformer architecture. The structure of the B-tagging transformer model is more complex than the engine anomaly detection model, with around 9,000 parameters, notably more than the MHA layers in the Engine Anomaly Detection Model. The increased complexity of this model helps capture the nuanced relationships in the input data, as the input vectors are more complicated than in the first dataset. The final layer is a SoftMax layer, which provides a probabilistic distribution across the output classes, in this case representing different types of jets. Like the previous model, the B-tagging model also employs residual connections to circumvent potential issues related to the vanishing gradient problem and to allow the learning process to build deeper models.

\subsection{Gravitational Wave Model}
The transformer architecture is suitable for time series data. 
Specifically, it can be useful for the classification of different sources of gravitational waves detected by Advanced LIGO~\cite{Aasi_2015} and Advanced VIRGO~\cite{Acernese_2015}, and KAGRA~\cite{KAGRA:2020tym}.
The challenge of signal detection at GW facilities is the presence of high-amplitude noise and the presence of glitches that can mimic a signal. 
A transformer-based classifier is trained to distinguish between signals and backgrounds, represented both by glitches and noise, and signals. 
We use the dataset collected by the LIGO detectors during the first half of the third observing run (O3a), which took place between 1st April 2019 and 1st Oct 2019. 
The background dataset has excess power glitches~\cite{Robinet:2020lbf} and known GW-events removed. We create the glitch dataset by using the Transient instrumental glitches (often of unknown origin) flagged by Omicron~\cite{Robinet:2020lbf} as having excess power.
The time-series data are downsampled from 16384\,Hz to 2048\,Hz. 
For signals simulated Binary Black Hole~(BBH) mergers and simulated sine-Gaussian~(SG) events are used. Simulated signals are injected on top of the real background to imitate as closely as possible real-life detection.
The detailed description of the signal simulation parameters can be found in~\cite{raikman2023gwak}.

The transformer model for the LIGO dataset is more complicated than other models, primarily due to its longer sequence length (100 time steps). This model also incorporates layer normalization and residual connections, further bolstering its analytical capabilities. Following the final MHA layer, the data goes through two dense layers for additional computation and transformation. The model's structure concludes with a final sigmoid layer, which curates the outputs into a probability distribution suitable for signal background classification. This model demonstrates impressive efficacy in identifying gravitational wave signals, as it achieves an Area Under the Receiver Operating Characteristics curve (AUC) of 97.8\%. 
Detailed comparisons of each model's performance in TensorFlow and \texttt{hls4ml} are presented in the following sections.

%% file: 06_performance.tex
\section{Performance, Resource and Latency Estimation}
\label{sec:performance}

Understanding the balance between performance, resource usage, and latency is fundamental when designing and implementing models on FPGAs. In this section, we delve into the nuances of our process, focusing on  the optimizations made and the results obtained.
Our process uses Vivado HLS 2019.2 and a Xilinx Ultra Scale FPGA VU13P. This large FPGA is used across all models to ensure a fair comparison of performance. In a significant deviation from conventional practices, our models don't use floating-point representations. Instead, we have quantized these representations and used fixed points in the FPGA inference. Initially, we deployed post-training quantization and then incorporated quantization-aware training. Additionally, we examined the trade-off between resource usage and the speed of model throughput, using the concept of the reuse factor for parallelization.
In the following subsections, we explore in detail the facets of post-training quantization, quantization-aware training, and parallelization.

\subsection{Quantization}
Quantization, specifically in the domain of machine learning, entails the process of diminishing the numerical precision of a model's parameters. Floating-point numbers, known for their expansive dynamic range, have usually been used in processing data in CPU or GPU. However, when transitioning to hardware implementations, like FPGA, fixed-point numbers are favored due to their reduced demand on memory and computational resources, making the computation much faster than with floating-point. For instance, consider an unsigned fixed-point number composed of 4 integer bits and 3 fractional bits. This numerical representation can store values ranging from 0 to 15.875, with a granularity, or step size, of 0.125. This ability to discretize values in such a manner, coupled with reduced resource demands, makes fixed-point numbers an good choice for FPGA-based machine learning implementations.

One method to convert from a floating-point range to a fixed-point range is post-training quantization (PTQ). In this approach, after the model has been trained with floating-point numbers, we convert it fixed point values within a hardware description language suitable for FPGA implementations using \texttt{hls4ml}. To enhance quantization quality, we also explored quantization-aware training (QAT). This technique trains the model while making it cognizant of its quantization constraints, often yielding superior accuracy compared to the PTQ approach. Our QAT implementation leverages the Qkeras package~\cite{coelho2021automatic}, extension of the Keras library tailored for quantization-aware training. The package equips neural network layers with quantization functionality. It's important to highlight, however, that the current version of Qkeras does not support certain layers, specifically MHA, SoftMax, and Layer Normalization. To address this gap, we enhanced the package by incorporating quantizers for each of these layers, thus achieving their quantized versions within the Qkeras environment.

In \texttt{hls4ml}, the bit precision for the fixed point can vary between layers, granting users control over precision. For the sake of testing and comparison, we kept the same precision across all layers. However, within \texttt{hls4ml}, there are different types of bits. For example, an accumulation type of bit usually has a larger integer bit width. We set this as a larger fixed number, 10 bits including the sign bit, and altered the fractional bit width for testing precision under different bit lengths.

\begin{figure}[!ht] 
    \centering
    \includegraphics[width=0.5\textwidth]{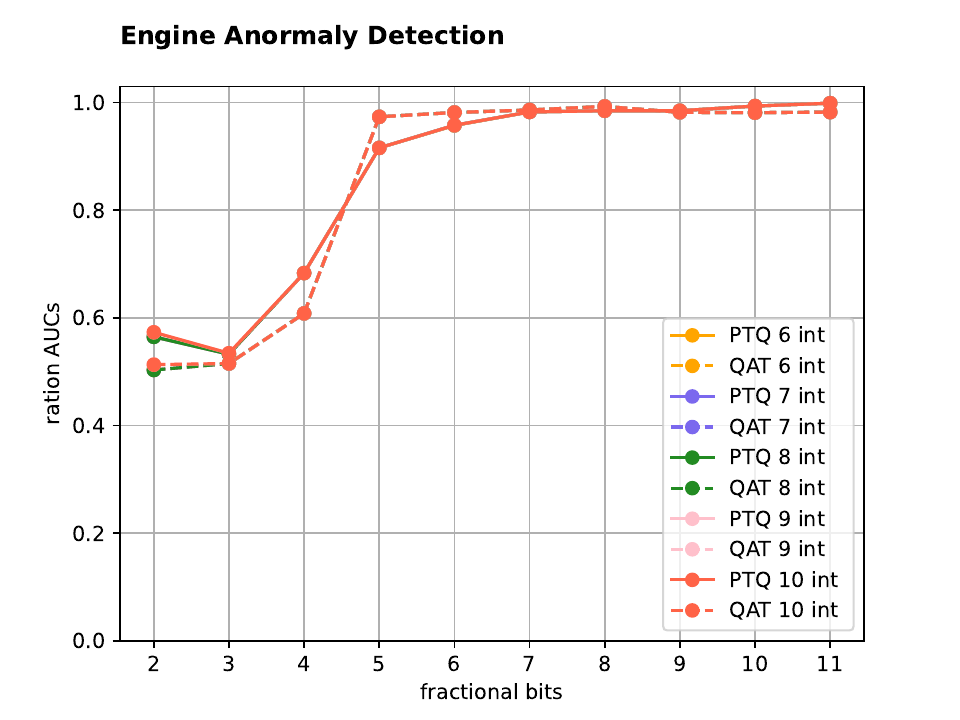} 
    \caption{The AUC plot of the car engine anomaly detection model. Most accuracy overlaps with each other.}
    \label{fig:auc_engine}
\end{figure}

\begin{figure}[!ht] 
    \centering
    \includegraphics[width=0.5\textwidth]{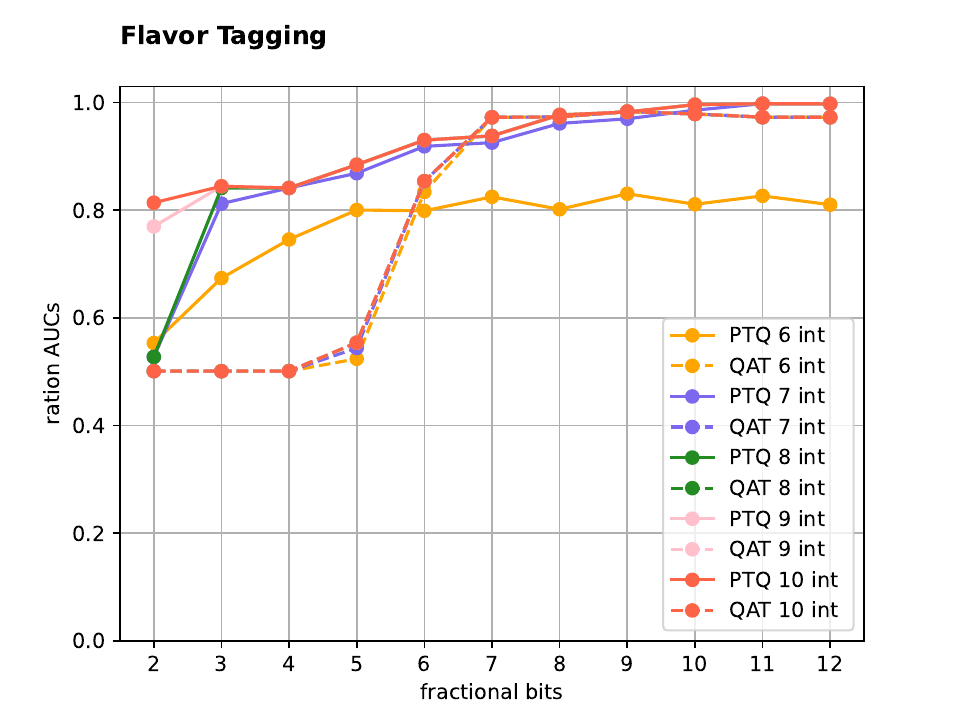} 
    \caption{The AUC plot of the B-tagging model. The QAT's result remains invalid at 0.5 until 6 fractional bits.}
    \label{fig:auc_btagging}
\end{figure}

\begin{figure}[!ht] 
    \centering
    \includegraphics[width=0.5\textwidth]{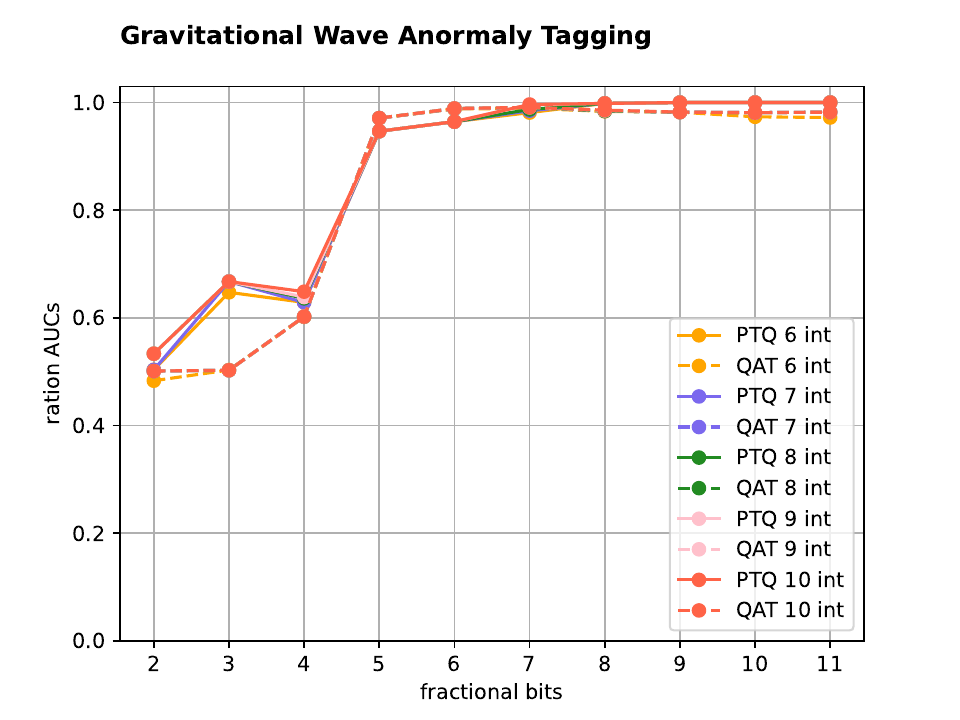} 
    \caption{The AUC plot of the gravitational wave anomaly detection model. Most accuracy overlaps with each other.}
    \label{fig:auc_gw}
\end{figure}

 The AUC metrics, for different bit widths, are graphically depicted in figure \ref{fig:auc_engine}, \ref{fig:auc_btagging}, and \ref{fig:auc_gw}. It is crucial to understand that these AUC metrics are derived from comparing the outputs of the Keras/QKeras model and the \texttt{hls4ml} model, rather than comparing the \texttt{hls4ml} output with the ground truth of the dataset, because we are primarily interested in the capability of \texttt{hls4ml} to replicate the output of the Keras and QKeras model accurately.

The plots provide us with insight into the relationship between bit width and model performance, as each model achieved its best performance, indicated by the AUC ratio, with a specific fractional bit width. After analysis, we deduced the optimal bit width for each model's implementation on the FPGA. For the engine anomaly detection model, we use 6 integer bits for both PTQ and QAT models; for the B-tagging model, it's 10 bits for the PTQ model and 6 bits for the QAT model; and for the gravitational wave model, it's 6 bits for both PTQ and QAT models. In later sections, we will use these settings as fixed values to evaluate the speed and resource usage of the model.

\subsection{Parallelization}
\label{subsec:Parallelization}
Parallelization is another key aspect of FPGA optimization in \texttt{hls4ml}, coordinated mainly through a parameter mechanism known as "reuse". This parameter represents the number of multiplication operations to be performed by each digital signal processing (DSP) block for a given matrix multiplication.

In cases where the reuse factor is set to 1, leading to a fully parallel configuration, each multiplication is handled independently by a separate DSP and can, therefore, occur concurrently. As we increase the reuse factor, the number of required DSPs decreases. However, this reduction comes at a cost: the latency and initiation interval of layer computations increases proportionally to the reuse.
To explore the implications of this trade-off, we synthesized all three benchmark models with varying values of the reuse factor and fractional bit precision. We quantified the results across multiple FPGA resource categories such as onboard FPGA memory (BRAM), DSPs, registers, and programmable logic elements like flip-flops (FFs) and lookup tables (LUTs).

In \texttt{hls4ml}, the synthesis of a model can follow one of two strategies: latency strategy, which aims to minimize latency, or resource strategy, which focuses on reducing resource utilization. The transformer architecture's complexity necessitates parallel processing of data across multiple time steps and feature dimensions. Consequently, we employed a layered strategy: smaller components like the dense layer, normalization layer, and SoftMax layer within the MHA block adhered to the latency strategy, thus producing outputs every cycle. On the other hand, the top level of the transformer model, due to its larger size, invariably followed the resource strategy. In this manner, the design is optimized for low resource utilization by leveraging the same hardware to execute operations across multiple stages.

\begin{figure}[!ht] 
    \centering
    \includegraphics[width=0.50\textwidth]{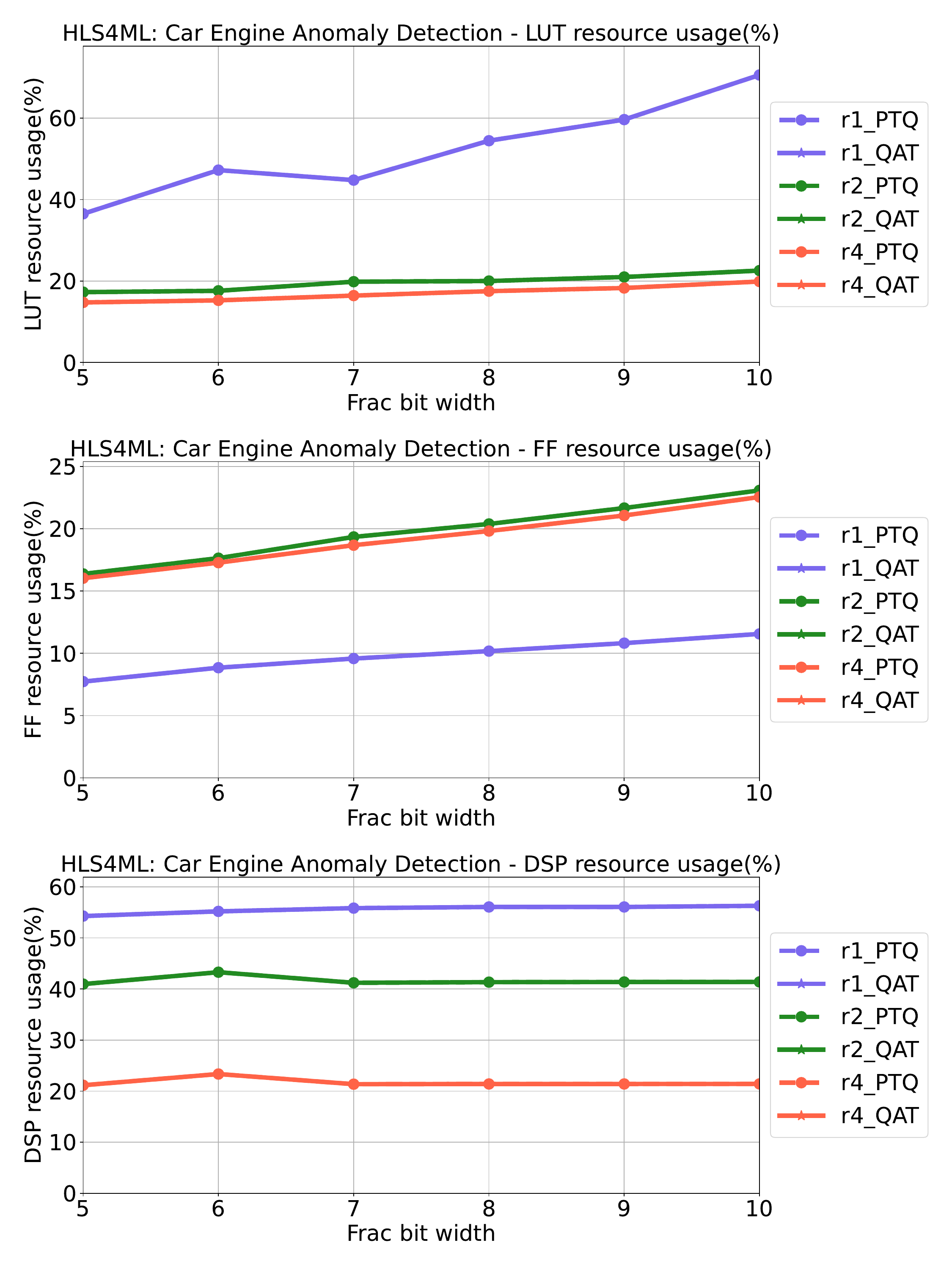} 
    \caption{The resource usage plot of the car engine anomaly detection model}
    \label{fig:resource_engine}
\end{figure}

\begin{figure}[!ht] 
    \centering
    \includegraphics[width=0.50\textwidth]{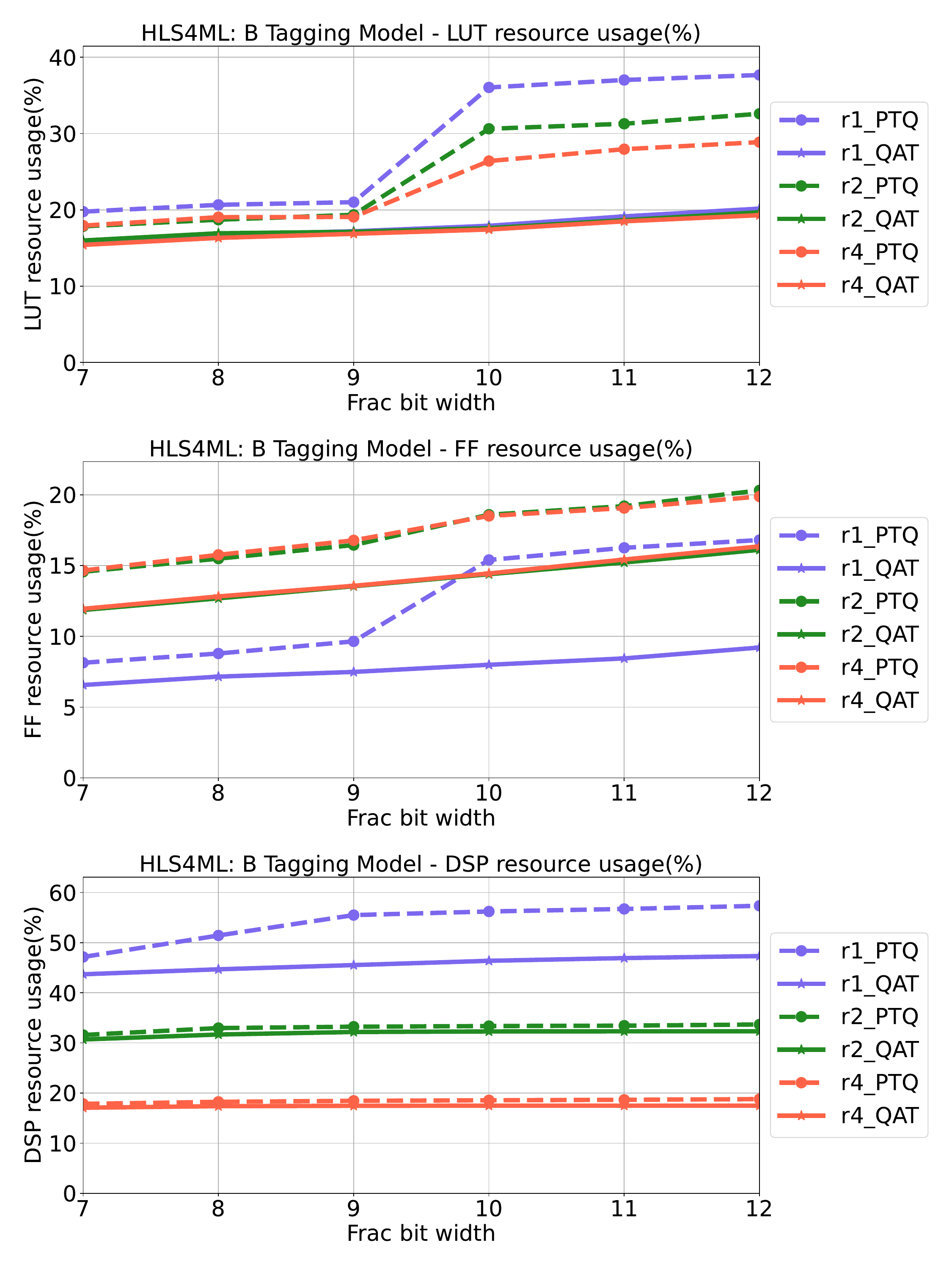} 
    \caption{The resource usage plot of the B tagging model}
    \label{fig:resource_btagging}
\end{figure}

\begin{figure}[!ht] 
    \centering
    \includegraphics[width=0.50\textwidth]{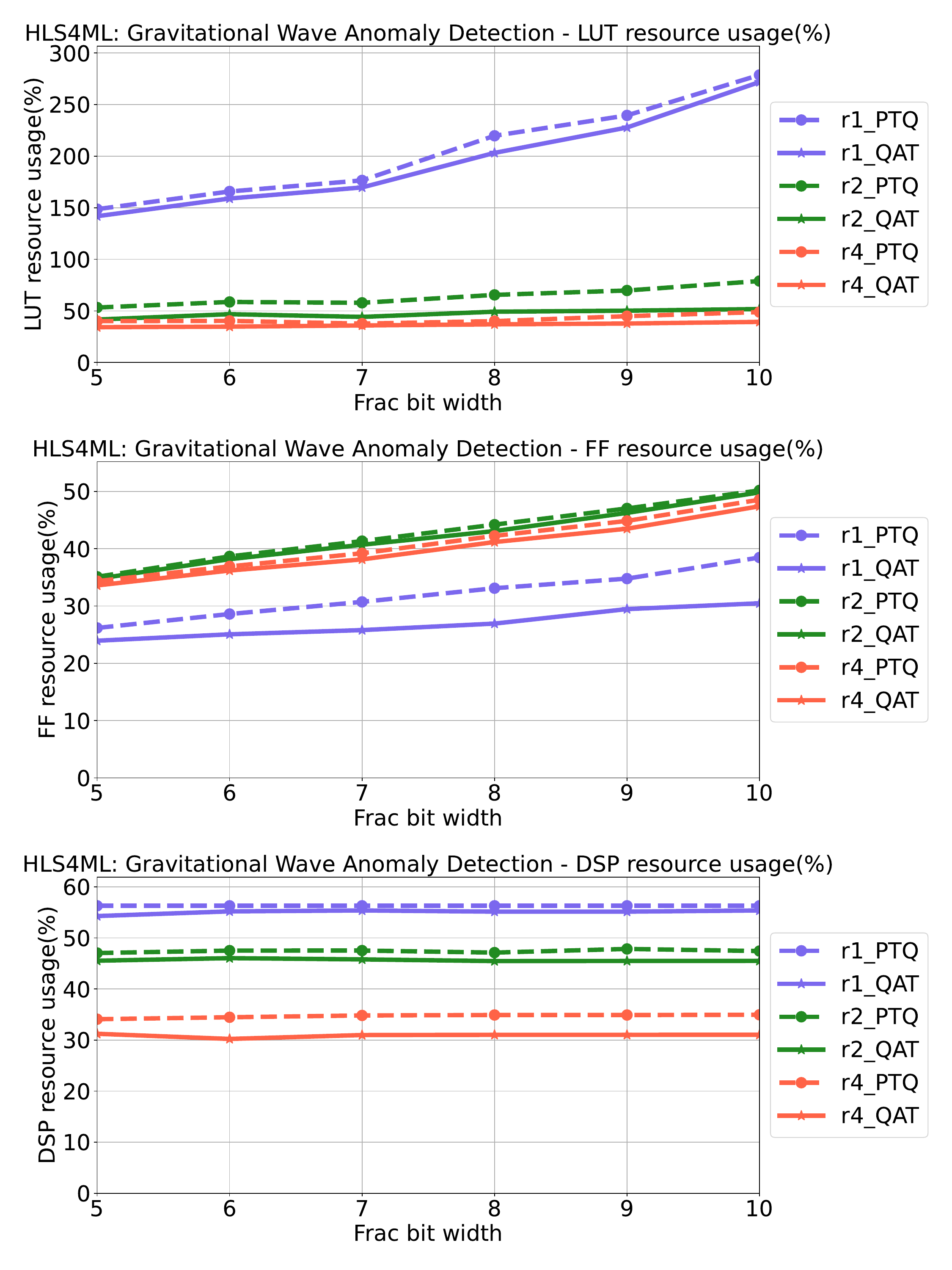} 
    \caption{The resource usage plot of the gravitational wave anomaly detection model}
    \label{fig:resource_gw}
\end{figure}

The minimum and maximum latencies for each model are cataloged in TABLE \ref{table:latency_engine}, \ref{table:latency_btagging}, and \ref{table:latency_gw}. Figures \ref{fig:resource_engine}, \ref{fig:resource_btagging}, and \ref{fig:resource_gw} depict the usage of DSPs, FFs, and LUTs respectively for each model across different reuse factor values. Here, the reuse factor values are represented as \(R\). All results are from Vivado mapped designs.

\begin{table}[ht]
\centering 
\caption{\label{table:latency_engine}Latency and Clock Period Analysis for Different Reuse Values of the Car Engine model}
\begin{tabular}{|c|c|c|c|c|c|}
\hline
\begin{tabular}[c]{@{}c@{}}Quantization\\ Type\end{tabular} & Reuse & clk (ns) & \begin{tabular}[c]{@{}c@{}}Interval\\ (cycle)\end{tabular} & \begin{tabular}[c]{@{}c@{}}Latency \\ (cycles)\end{tabular} & \begin{tabular}[c]{@{}c@{}}Latency \\ (us)\end{tabular} \\ \hline
\multirow{3}{*}{PTQ}                                        & R1    & 7.423    & 119                                                        & 257                                                         & 1.908                                                   \\ \cline{2-6} 
                                                            & R2    & 4.367    & 218                                                        & 456                                                         & 2.280                                                   \\ \cline{2-6} 
                                                            & R4    & 4.367    & 318                                                        & 756                                                         & 3.780                                                   \\ \hline
\multirow{3}{*}{QAT}                                        & R1    & 7.423    & 119                                                        & 257                                                         & 1.908                                                   \\ \cline{2-6} 
                                                            & R2    & 4.367    & 218                                                        & 456                                                         & 2.280                                                   \\ \cline{2-6} 
                                                            & R4    & 4.367    & 318                                                        & 756                                                         & 3.780                                                   \\ \hline
\end{tabular}
\end{table}

\begin{table}[!ht]
\centering
\caption{\label{table:latency_btagging}Latency and Clock Period Analysis for Different Reuse Values of the  B-tagging model}
\begin{tabular}{|c|c|c|c|c|c|}
\hline
\begin{tabular}[c]{@{}c@{}}Quantization\\ Type\end{tabular} & Reuse & clk (ns) & \begin{tabular}[c]{@{}c@{}}Interval\\ (cycle)\end{tabular} & \begin{tabular}[c]{@{}c@{}}Latency \\ (cycles)\end{tabular} & \begin{tabular}[c]{@{}c@{}}Latency \\ (us)\end{tabular} \\ \hline
\multirow{3}{*}{PTQ}                                        & R1    & 6.577    & 49                                                         & 269                                                         & 2.077                                                   \\ \cline{2-6} 
                                                            & R2    & 6.215    & 65                                                         & 449                                                         & 3.467                                                   \\ \cline{2-6} 
                                                            & R4    & 4.723    & 100                                                        & 768                                                         & 5.853                                                   \\ \hline
\multirow{3}{*}{QAT}                                        & R1    & 6.568    & 48                                                         & 266                                                         & 2.055                                                   \\ \cline{2-6} 
                                                            & R2    & 6.210    & 63                                                         & 445                                                         & 3.440                                                   \\ \cline{2-6} 
                                                            & R4    & 4.722    & 99                                                         & 767                                                         & 5.848                                                   \\ \hline
\end{tabular}
\end{table}

\begin{table}[!ht]
\centering
\caption{\label{table:latency_gw}Latency and Clock Period Analysis for Different Reuse Values of the Gravitational Wave model}
\begin{tabular}{|c|c|c|c|c|c|}
\hline
\begin{tabular}[c]{@{}c@{}}Quantization\\ Type\end{tabular} & Reuse & clk (ns) & \begin{tabular}[c]{@{}c@{}}Interval\\ (cycle)\end{tabular} & \begin{tabular}[c]{@{}c@{}}Latency \\ (cycles)\end{tabular} & \begin{tabular}[c]{@{}c@{}}Latency \\ (us)\end{tabular} \\ \hline
\multirow{3}{*}{PTQ}                                        & R1    & 6.577    & 212                                                        & 537                                                         & 3.532                                                   \\ \cline{2-6} 
                                                            & R2    & 6.215    & 412                                                        & 1035                                                        & 6.433                                                   \\ \cline{2-6} 
                                                            & R4    & 4.723    & 612                                                        & 1835                                                        & 9.175                                                   \\ \hline
\multirow{3}{*}{QAT}                                        & R1    & 6.577    & 210                                                        & 532                                                         & 3.499                                                   \\ \cline{2-6} 
                                                            & R2    & 6.215    & 411                                                        & 1033                                                        & 6.420                                                   \\ \cline{2-6} 
                                                            & R4    & 4.723    & 611                                                        & 1834                                                        & 9.170                                                   \\ \hline
\end{tabular}
\end{table}

A noticeable trend across all resources is their general increase with decreasing values of \(R\) and increased precision. For FFs and LUTs, this increase is approximately linear, whereas DSP utilization remains consistent until the precision surpasses the DSP input width. Upon exceeding this threshold, an additional DSP is employed for computations. Another fact we can discover from the plot is that latency shows the opposite trend to FF and LUT utilization. Thus, similar to other architectures supported by \texttt{hls4ml} (like CNNs or RNNs), reuse can be manipulated to reduce FF and LUT counts while incurring a latency penalty. This flexibility is critical for users to tune resource usage and latency, allowing synthetic designs to meet specified requirements.

Furthermore, the reuse factor influences how array partitioning and memory storage are executed in the transformer model. In conventional \texttt{hls4ml} approaches, all values are stored in registers without BRAM utilization. However, in the transformer architecture, not all values need to be accessed in parallel. Hence, we also used the reuse factor to partition array values and store them in BRAM, leading to more efficient memory usage.

%% file: 07_conclusion.tex
\section{Conclusion}
\label{sec:conclusion}

This paper introduces an effective way to apply transformer models using \texttt{hls4ml}. Our work opens up new opportunities for various fields, particularly in physics research, where quick and accurate computations are critical. Beyond that, the approach we've developed can be applied wherever transformer models are used, extending the benefits of our work to many other areas.

The approach we've developed has several significant benefits. The pipeline structure we use speeds up the inference time, making computations faster. By using a combination of post-training and quantization-aware training strategies, our models achieve high accuracy with less resource usage compared to the typical floating-point precision approach. This balance between speed, accuracy, and efficient use of resources makes our method a powerful tool for any application using FPGA.

Looking ahead, there are many exciting possibilities to build on this work. We could add masking ability to the MHA layer to make transformer models more flexible. We could also develop a version of our transformer implementation that uses sparse computations for the dense layer, a growing trend in deep learning that can save resources. By continuing to innovate in this way, we aim to make \texttt{hls4ml} an even more effective tool for deploying deep learning.

%% file: transformer_hls4ml.bbl
\begin{thebibliography}{10}
\providecommand{\url}[1]{#1}
\csname url@samestyle\endcsname
\providecommand{\newblock}{\relax}
\providecommand{\bibinfo}[2]{#2}
\providecommand{\BIBentrySTDinterwordspacing}{\spaceskip=0pt\relax}
\providecommand{\BIBentryALTinterwordstretchfactor}{4}
\providecommand{\BIBentryALTinterwordspacing}{\spaceskip=\fontdimen2\font plus
\BIBentryALTinterwordstretchfactor\fontdimen3\font minus
  \fontdimen4\font\relax}
\providecommand{\BIBforeignlanguage}[2]{{%
\expandafter\ifx\csname l@#1\endcsname\relax
\typeout{** WARNING: IEEEtran.bst: No hyphenation pattern has been}%
\typeout{** loaded for the language `#1'. Using the pattern for}%
\typeout{** the default language instead.}%
\else
\language=\csname l@#1\endcsname
\fi
#2}}
\providecommand{\BIBdecl}{\relax}
\BIBdecl

\bibitem{LHC_2008}
L.~Evans and P.~Bryant, ``{LHC} machine,'' \emph{JINST}, vol.~3, pp.
  S08\,001--S08\,001, aug 2008. \doi{10.1088/1748-0221/3/08/s08001}

\bibitem{Aasi_2015}
\BIBentryALTinterwordspacing
T.~L.~S. Collaboration and J.~A. et~al., ``Advanced ligo,'' \emph{Classical and
  Quantum Gravity}, vol.~32, p. 074001, mar 2015.
  \doi{10.1088/0264-9381/32/7/074001}. [Online]. Available:
  \url{https://dx.doi.org/10.1088/0264-9381/32/7/074001}
\BIBentrySTDinterwordspacing

\bibitem{Acernese_2015}
\BIBentryALTinterwordspacing
F.~A. et~al., ``Advanced virgo: a second-generation interferometric
  gravitational wave detector,'' \emph{Classical and Quantum Gravity}, vol.~32,
  p. 024001, dec 2014. \doi{10.1088/0264-9381/32/2/024001}. [Online].
  Available: \url{https://dx.doi.org/10.1088/0264-9381/32/2/024001}
\BIBentrySTDinterwordspacing

\bibitem{KAGRA:2020tym}
{KAGRA} Collaboration, ``{Overview of KAGRA: Detector design and construction
  history},'' \emph{PTEP}, vol. 2021, p. 05A101, 2021.
  \doi{10.1093/ptep/ptaa125}.
  \href{http://www.arXiv.org/abs/2005.05574}{arXiv:2005.05574}

\bibitem{NIPS2017_3f5ee243}
\BIBentryALTinterwordspacing
A.~Vaswani, N.~Shazeer, N.~Parmar, J.~Uszkoreit, L.~Jones, A.~N. Gomez, L.~u.
  Kaiser, and I.~Polosukhin, ``Attention is all you need,'' in \emph{Advances
  in Neural Information Processing Systems}, I.~Guyon, U.~V. Luxburg,
  S.~Bengio, H.~Wallach, R.~Fergus, S.~Vishwanathan, and R.~Garnett, Eds.,
  vol.~30.\hskip 1em plus 0.5em minus 0.4em\relax Curran Associates, Inc.,
  2017. [Online]. Available:
  \url{https://proceedings.neurips.cc/paper_files/paper/2017/file/3f5ee243547dee91fbd053c1c4a845aa-Paper.pdf}
\BIBentrySTDinterwordspacing

\bibitem{Li_2020_ISLPED}
\BIBentryALTinterwordspacing
B.~Li, S.~Pandey, H.~Fang, Y.~Lyv, J.~Li, J.~Chen, M.~Xie, L.~Wan, H.~Liu, and
  C.~Ding, ``Ftrans: energy-efficient acceleration of transformers using
  fpga,'' ser. ISLPED '20.\hskip 1em plus 0.5em minus 0.4em\relax New York, NY,
  USA: Association for Computing Machinery, 2020, p. 175–180.
  \doi{10.1145/3370748.3406567}. [Online]. Available:
  \url{https://doi.org/10.1145/3370748.3406567}
\BIBentrySTDinterwordspacing

\bibitem{Wojcicki_2022_ICFPT}
F.~Wojcicki, Z.~Que, A.~D. Tapper, and W.~Luk, ``Accelerating transformer
  neural networks on fpgas for high energy physics experiments,'' in \emph{2022
  International Conference on Field-Programmable Technology (ICFPT)}, 2022, pp.
  1--8. \doi{10.1109/ICFPT56656.2022.9974463}

\bibitem{Tzanos_2022_PACET}
G.~Tzanos, C.~Kachris, and D.~Soudris, ``Hardware acceleration of transformer
  networks using fpgas,'' in \emph{2022 Panhellenic Conference on Electronics
  Telecommunications (PACET)}, 2022, pp. 1--5.
  \doi{10.1109/PACET56979.2022.9976354}

\bibitem{Han_2023_FPL}
\BIBentryALTinterwordspacing
Y.~Han and Q.~Liu, ``Hpta: A high performance transformer accelerator based on
  fpga,'' in \emph{2023 33rd International Conference on Field-Programmable
  Logic and Applications (FPL)}.\hskip 1em plus 0.5em minus 0.4em\relax Los
  Alamitos, CA, USA: IEEE Computer Society, sep 2023, pp. 27--33.
  \doi{10.1109/FPL60245.2023.00012}. [Online]. Available:
  \url{https://doi.ieeecomputersociety.org/10.1109/FPL60245.2023.00012}
\BIBentrySTDinterwordspacing

\bibitem{Duarte:2018ite}
J.~Duarte \emph{et~al.}, ``{Fast inference of deep neural networks in FPGAs for
  particle physics},'' \emph{JINST}, vol.~13, p. P07027, 2018.
  \doi{10.1088/1748-0221/13/07/P07027}.
  \href{http://www.arXiv.org/abs/1804.06913}{arXiv:1804.06913}

\bibitem{CNN_hls4ml}
T.~Aarrestad \emph{et~al.}, ``Fast convolutional neural networks on fpgas with
  hls4ml,'' \emph{Machine Learning: Science and Technology}, vol.~2, p. 045015,
  jul 2021.

\bibitem{RNN_hls4ml}
E.~E. Khoda \emph{et~al.}, ``{Ultra-low latency recurrent neural network
  inference on FPGAs for physics applications with hls4ml},'' \emph{Mach.
  Learn. Sci. Tech.}, vol.~4, p. 025004, 2023. \doi{10.1088/2632-2153/acc0d7}.
  \href{http://www.arXiv.org/abs/2207.00559}{arXiv:2207.00559}

\bibitem{RNN_GW_hls4ml}
Z.~Que \emph{et~al.}, ``{Accelerating Recurrent Neural Networks for
  Gravitational Wave Experiments},'' in \emph{{32nd IEEE International
  Conference on Application-specific Systems, Architectures and Processors}}, 6
  2021. \doi{10.1109/ASAP52443.2021.00025}.
  \href{http://www.arXiv.org/abs/2106.14089}{arXiv:2106.14089}

\bibitem{GNN_hls4ml}
A.~Elabd \emph{et~al.}, ``{Graph Neural Networks for Charged Particle Tracking
  on FPGAs},'' \emph{Front. Big Data}, vol.~5, p. 828666, 2022.
  \doi{10.3389/fdata.2022.828666}.
  \href{http://www.arXiv.org/abs/2112.02048}{arXiv:2112.02048}

\bibitem{Peng_ISQED_2021}
H.~Peng, S.~Huang, T.~Geng, A.~Li, W.~Jiang, H.~Liu, S.~Wang, and C.~Ding,
  ``Accelerating transformer-based deep learning models on fpgas using column
  balanced block pruning,'' in \emph{2021 22nd International Symposium on
  Quality Electronic Design (ISQED)}, 2021, pp. 142--148.
  \doi{10.1109/ISQED51717.2021.9424344}

\bibitem{hong2022dfx}
S.~Hong, S.~Moon, J.~Kim, S.~Lee, M.~Kim, D.~Lee, and J.-Y. Kim, ``Dfx: A
  low-latency multi-fpga appliance for accelerating transformer-based text
  generation,'' 2022.

\bibitem{UCRArchive2018}
H.~A. Dau \emph{et~al.}, ``The ucr time series classification archive,''
  October 2018, \url{https://www.cs.ucr.edu/~eamonn/time_series_data_2018/}.

\bibitem{ftag_dataset}
``Mc: Ttbar sample from the cms hep tutorial,''
  \url{http://opendata.cern.ch/record/204}.

\bibitem{Robinet:2020lbf}
F.~Robinet, N.~Arnaud, N.~Leroy, A.~Lundgren, D.~Macleod, and J.~McIver,
  ``{Omicron: a tool to characterize transient noise in gravitational-wave
  detectors},'' \emph{SoftwareX}, vol.~12, p. 100620, 2020.
  \doi{10.1016/j.softx.2020.100620}.
  \href{http://www.arXiv.org/abs/2007.11374}{arXiv:2007.11374}

\bibitem{raikman2023gwak}
R.~Raikman \emph{et~al.}, ``Gwak: Gravitational-wave anomalous knowledge with
  recurrent autoencoders,'' 2023.

\bibitem{coelho2021automatic}
C.~N. C.~J. au2, A.~Kuusela, S.~Li, H.~Zhuang, T.~Aarrestad, V.~Loncar,
  J.~Ngadiuba, M.~Pierini, A.~A. Pol, and S.~Summers, ``Automatic heterogeneous
  quantization of deep neural networks for low-latency inference on the edge
  for particle detectors,'' 2021.

\end{thebibliography}
